\documentclass[letterpaper, 10 pt, conference]{ieeeconf}  % Comment this line out if you need a4paper

\IEEEoverridecommandlockouts                              % This command is only needed if 
\usepackage{epsfig} % for postscript graphics files
\usepackage{xcolor}
\usepackage{bm}
\usepackage{amsmath,amssymb,array,balance,booktabs,cite,changes,color,enumerate,float,graphicx,hyperref,multicol,multirow,siunitx,times,subcaption,mathtools,colortbl,}

\usepackage{xcolor}
\newcommand{\cmt}[1]{}

\newcommand{\jhc}[1]{\textcolor{violet}{{#1}}} % for added contents that is not revised

\long\def\ignorethis#1{}

\newcommand{\pctab}{\hspace{0.2in}}

\definecolor{myblue}{rgb}{0.44, 0.65, 0.82}

\renewcommand\vec{\mathbf}
\DeclareMathOperator*{\argmin}{arg\,min}

\title{\LARGE \bf
ARMP: Autoregressive Motion Planning for Quadruped Locomotion and Navigation in Complex Indoor Environments}

\author{Jeonghwan Kim, Tianyu Li, Sehoon Ha
\thanks{Georgia Institute of Technology, Atlanta, GA, 30308, USA}
\thanks{{\tt \small jkim3662@gatech.edu, tli471@gatech.edu, sehoonha@gatech.edu}}%
}

\begin{document}

\maketitle
\thispagestyle{empty}
\pagestyle{empty}

% !TeX root = ./main.tex
\begin{abstract}
Generating natural and physically feasible motions for legged robots has been a challenging problem due to its complex dynamics. In this work, we introduce a novel learning-based framework of autoregressive motion planner (ARMP) for quadruped locomotion and navigation. Our method can generate motion plans with an arbitrary length in an autoregressive fashion, unlike most offline trajectory optimization algorithms for a fixed trajectory length. To this end, we first construct the motion library by solving a dense set of trajectory optimization problems for diverse scenarios and parameter settings.
Then we learn the motion manifold from the dataset in a supervised learning fashion. We show that the proposed ARMP can generate physically plausible motions for various tasks and situations. We also showcase that our method can be successfully integrated with the recent robot navigation frameworks as a low-level controller and unleash the full capability of legged robots for complex indoor navigation.

\end{abstract}

% !TeX root = ./main.tex

\section{Introduction}
% 1. Motion planning is a tough problem for high-dof robots, in many different applications
Generating effective and plausible motions for robotic creatures has been one of the most important topics in various autonomous robotic missions, including manipulation, locomotion, and navigation. For instance, a manipulator in cluttered environments needs to find a collision-free path to reach the target object. A legged robot in challenging environments needs to carefully its footsteps to approach the destination while maintaining its balance. However, motion planning becomes more and more challenging for higher-dimensional robots, such as quadruped or humanoid robots, because the search space size is exponential to the number of degrees of freedom (DoFs). Researchers have proposed many approaches, such as simplified models, to find an optimal trajectory in such high-dimensional spaces, but it is still an ongoing research topic in the field of robotics.

\begin{figure}[t]
\centering
    \begin{subfigure}[b]{0.45\textwidth}
        \centering
        \includegraphics[width=0.95\textwidth]{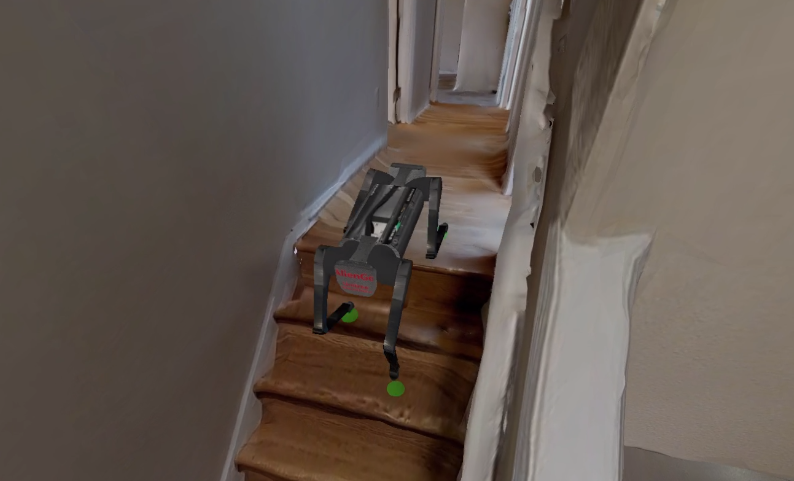}
    \end{subfigure}
\caption{\small{We present a framework for learning an auto-regressive motion planner (ARMP) to generate physically plausible motion plans for various tasks and environments. Here we show using ARMP as the underlying motion planner for an indoor navigation task.}}
\label{fig: profile}
\vspace{-10pt}
\end{figure}

% 2. This becomes limitation in navigation. Simple models (teleported). Bridging gap.
This challenge of motion planning is also another major hurdle in developing a visual navigation controller for high-dimensional robots, such as 12 DoF quadrupedal robots. Many navigation simulatiors~\cite{xia2018gibson, szot2021habitat} abstract away robotic agents' kinematic and dynamic features by adopting a simple state representation that consists of the global position and orientation, ignoring the transition dynamics. Due to this abstraction, high-level navigation planners cannot utilize the full capability of the quadrupedal robot, such as jumping, walking over obstacles, or climbing stairs. However, it is still not straightforward how we can generate motions effectively due to its expensive computational cost.

% 3. (Literature) quadruped offline vs. online
Over the decades, researchers have proposed various approaches for planning kinematic or dynamic trajectories. One popular approach is trajectory optimization (TO) which solves an open-loop motion plan offline. Offline trajectory optimization can also be formulated in multiple ways: for instance, a collocation method discretizes the trajectory over time and optimizes it with respect to the given task objectives and dynamics constraints. However, offline planning often requires specifying a fixed time horizon, which may not be suitable for navigation tasks with either unknown or infinite time duration. On the other hand, a shooting method leverages simulation to generate trajectories in a forward direction. However, designing such a robust dynamic controller is very challenging despite the recent advances in visual locomotion \cite{kumar2021rma, yu2021visual, jenelten2020perceptive, grandia2022perceptive}. In addition, physics simulation for quadrupedal locomotion typically requires $500$ to $1,000$ Hz simulation, which would be too expensive to train a visual navigation agent.

\begin{figure*}[ht]
\centerline{\includegraphics[width=0.9\linewidth]{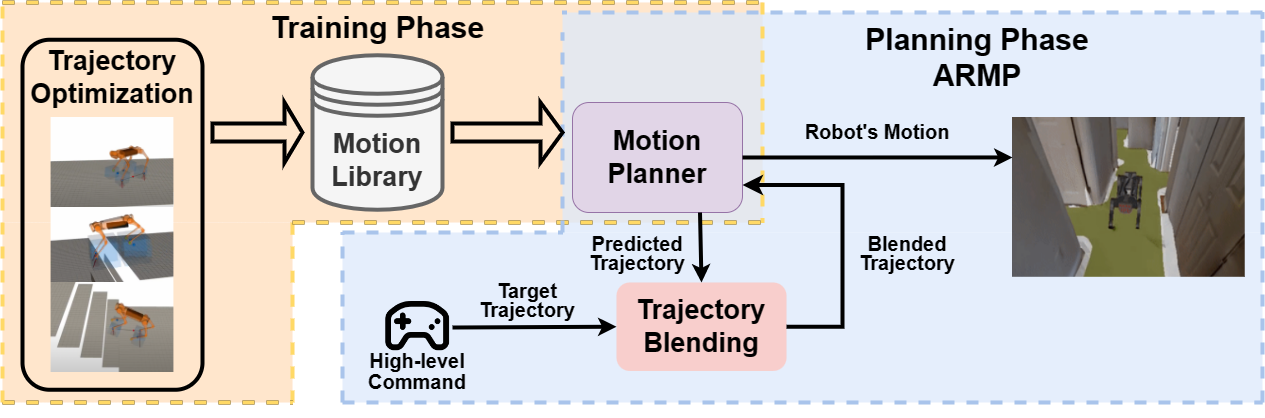}}
\caption{\small{Overview of the framework. In the training phase, a motion library is constructed by conducting trajectory optimization in different settings. In the planning phase, the motion planner leverages an autoregressive method by taking a blended desired trajectory as part of the input for predicting the robot's motion.}}
\label{fig: Overview}
\vspace{-10pt}
\end{figure*}

% 4. We present ARMP... We take inspiration from...
We present a novel autoregressive motion planner (ARMP), which is an efficient approach to synthesizing quadrupedal motion plans from the database. To ensure the physical plausibility of our motion planner, we create a dense set of motion trajectories over various tasks and environments using the traditional offline motion planning algorithm \cite{winkler2018gait}. Then we train a neural network that learns to sequentially generates motion based on previous frames and user control. We take inspiration from the recent advances in character control in computer animation \cite{holden2017phase, zhang2018mode, starke2019neural} and adopt a mixture-of-expert network architecture.%%\ty{Still purple here} \jhc{Their autoregressive architecture enables the planner to easily take account of environment and user control, which is essential for locomotion under complex environments.}

% 5. Our results
We showcase that the proposed ARMP framework can efficiently generate natural and feasible motions of the Aliengo quadrupedal robot \cite{unitree}. It can automatically adapt its gaits to overcome various terrains, including slopes, stairs, and flat terrain with obstacles that require the robot to jump over. We ensure the physical validity of the generated trajectories by reproducing the trajectories in a physics simulator~\cite{raisim}. Because it runs at real-time 
and does not require any internal optimization, ARMP can be easily combined with the existing embodied AI and autonomous navigation framework, Habitat \cite{habitat19iccv, szot2021habitat}, which can be potentially used for future learning of quadrupedal robot navigation skills.
\section{Related Works}
%Why do we have to use our method for this particular problem?
\subsection{Legged Robot Navigation}

Our work is motivated by autonomous robotic navigation, which is the problem of finding a valid path to reach a target located in an environment. Although promising results have been presented for wheel-based robots~\cite{chaplot2020learning, kadian2020sim2real, hirose2022exaug, yokoyama2021success}, the navigation of legged robots is still at a premature stage without considering joint-level planning. Therefore, legged robot navigation is usually tackled using a hierarchical structure where the high-level controller generates a command variable for a low-level controller~\cite{kareer2022vinl, truong2021learning} who runs at a higher frequency. The low-level controller can be either expert designed ~\cite{kareer2022vinl,truong2021learning,sorokin2022learning} or learned using reinforcement learning~\cite{fu2022coupling, jain2020pixels}. However, most of the previous works only considers environments with flat surfaces, which do not fully exploit the potential of the legged robot. Although a recent work\cite{kareer2022vinl} addresses the legged robot navigation with small obstacles scattered on the surface, the obstacles are only treated as disturbances. In this work, our goal is to generate a feasible motion plan for a legged robot in realistic indoor environments that include stairs and obstacles.

% I think TO equals MP in quadrupedal robotics.
\subsection{Motion Planning for legged robots}
Motion planning is a research domain that addresses the problem of generating robot motion trajectories for given tasks, which is often approached by trajectory optimization~\cite{winkler2018gait, carius2019trajectory, cebe2021online}. Due to the nonlinearity and high dimensionality of the nature of the robot's dynamics, multiple simplifications are applied to solve the problem, such as a single rigid body model~\cite{ding2019real, di2018dynamic} or inverted pendulum models \cite{green2021learning}. Winkler et al.~\cite{winkler2018gait} reduces the problem space by by using Hermite polynomials for parameterizing the kinodynamic trajectories, where he simplifies further by removing the cost function and treats trajectory optimization as a feasibility problem. Even with these simplifications, solving trajectory optimization problems with long-horizon and complex terrains is still relatively slow and unstable, where the quality and optimization time is highly sensitive to the initial condition. Recent research leverage machine learning methods to accelerate the motion planning process. Mansard et al. \cite{mansard2018using} use deep neural networks to memorize the previous optimized solutions and use them to warm-start the next optimization. Kurtz et al. \cite{kurtz2022mini} train a neural network to plan an entire robot trajectory given the initial state of a robot. The key difference between our method and \cite{kurtz2022mini} is that our generated motion can be modified according to a high-level control variable while \cite{kurtz2022mini} creates a fixed trajectory based only on the initial state of the robot.

\subsection{Kinematic Animation}
Various approaches have been discussed to generate plausible motions from the dataset. One classical method is to construct a motion graph that links different motion clips manually \cite{rose1998verbs} or automatically \cite{kovar2008motion}. The main drawback of these motion graph methods is the inherent discreteness of their underlying mechanisms, where transitions only happen at the end of short motion segments. On the other hand, principle component analysis~(PCA) reduces the dimension of the motion data and produces full-body motions by adjusting variables in a continuous low-dimensional space \cite{howe1999bayesian, safonova2004synthesizing}. Yet, PCA-based methods have difficulty dealing with a large amount of motion data. Recent works have started to leverage deep neural networks and machine learning for motion synthesis. Phase-Functioned Neural Network (PFNN)~\cite{holden2017phase} learns an effective motion controller using its novel network architecture, which conditions its weights on a phase variable. Model-adaptive Neural Network~\cite{zhang2018mode} generates realistic animations of a quadrupedal character by introducing a gated network architecture. Starke and the colleague~\cite{starke2020local} introduce a technique to automatically extract local motion phase signals from the main joints of the skeleton. The work of Li et al.~\cite{li2022ganimator} adopts a generative model to synthesize new motion sequences from a single input clip.

\section{Auto-regressive Motion Planner}

We present Auto-regressive Motion Planner~(ARMP), a framework for generating physically plausible motions that can be used for legged robot navigation tasks in indoor environments. An overview of our method is presented in Fig. \ref{fig: Overview}. This framework consists of three components: (1) a motion library that contains a set of physically valid motions, (2) a motion planner that predicts the next state from the current state and the high-level commands, and (3) a trajectory blender that smoothes the target trajectory with the future trajectory predicted by the planner module. The details of each part are provided by the following subsections.

\subsection{Constructing Motion Library Via Trajectory Optimization } \label{sec:data_generation}

We construct a motion library $\mathcal{X}$ for training ARMP by collecting a set of physically feasible trajectories $\{\bm\tau^1, \bm\tau^2, \cdots, \bm\tau^N\}$. Each trajectory $\bm\tau^i$ is defined as a set of states $\{\mathbf{x}_1, \mathbf{x}_2, \cdots, \mathbf{x}_{T_i}\}$, where $T_i$ is the trajectory length which can be varied for each motion.  A more detailed definition of the state vector will be described in the following section.

In theory, our method is agnostic to how we generate these trajectories, so different methods can be applied for producing motion trajectories. For instance, we can generate trajectories by rolling out expert-designed controllers. However, designing controllers that can perform a large range of motions is not straightforward and requires significant expertise and human labor. Alternatively, reinforcement learning methods offer an automated way of motion generation. However, these methods suffer from issues such as sample inefficiency or unnatural result motions. In this work, we leverage the trajectory optimization method for generating motions. Compared to manually designing controllers and reinforcement learning, trajectory optimization can produce reliable motion trajectories without considerable engineering effort.  

Our trajectory optimization is based on TOWR~\cite{winkler2018gait}. We use a Single Rigid Body Dynamics(SRBD) model, where the state of a robot is simplified into root's position $r(t)$, orientation $w(t)$, and four end effector positions $p_j(t)$. This model assumes that the mass of the legs is negligible to that of the body, ignoring momentum produced by joint velocities and fixing the robot's inertia to the one in nominal configuration. Instead of optimizing the joint position and joint torques directly, we optimize the end-effector position $p_j(t)$ and ground reaction force $f_j(t)$ so that our optimization variables lie in Cartesian coordinates. To enforce a feasible kinematic structure, we constrain our end-effector position to a cuboid whose center is at each end-effector’s nominal position. We additionally apply friction pyramid constraints to prevent slipping. We enforce our force to be always pushing when in contact, and set it to 0 when it is in a swing phase. The feet in the stance phase should have a z-position equal to the terrain height and the feet in the swing phase should have a z-position greater than the terrain height. Overall, the optimization equations are as follow:
\begin{equation*}
\begin{aligned}  
    \bar{p}_j(t), \bar{f}_j(t) &= \argmin C(p_j(t), f_j(t)) \\
    \text{s.t.} \ M \ddot{r}(t) &= \sum_{j=1}^{4} f_j(t) - M g \\
    I \dot{w}(t) + w(t) \times I w(t) &= \sum_{j=1}^{4} f_j(t) \times (r(t) - p_j(t)) \\
    p_i &\in \text{Cuboid}_i(r(t))\\ % ee in cuboid
    (p_i)_z &\geq \text{TerrainHeight}(p_i)\\ % above terrain
    0 \leq (f_i)_z &\leq \mu \max\{|(f_i)_x|, |(f_i)_y|\}\\ % friction pyramid with pushing
    f_i &= 0 \text{ for swing foot}\\ % swing
    (p_i)_z &= \text{TerrainHeight}(p_i) \text{ fot stance foot.}\\ % on terrain on stance
\end{aligned}
\end{equation*}
Here, $g$ denotes gravity constant and the robot's mass $M$ and inertia in body frame $I$ are assumed to be constant over time.
Please refer to the paper~\cite{winkler2018gait} for the detailed description of the constraints

The state($r(t), p_i(t)$) and force($f_i(t)$) trajectories are parameterized using Hermite polynomials and optimized via IPOPT~\cite{ifopt} interfaced by IFOPT~\cite{wachter2006implementation}.  

% \subsection{Motion Planner Design and Learning}
\subsection{Learning-based Motion Planner Design}
Our autoregressive motion planner aims to predict a feasible next state from the current state and the user input based on the constructed motion library. We represent such an autoregressive motion planner as a neural network and train it via supervised learning. 
The input \jhc{to the network} at $i$-th time step is composed of:
\begin{itemize}
    \item robot state: $\mathbf{x}_i = \{ \mathbf{r}_i,\dot{\mathbf{r}_i}, \mathbf{w}_i, \mathbf{e}_i, \dot{\mathbf{e}_i}\}$,
    \item surrounding robot root trajectory: $\mathbf{t}_i = \{\mathbf{t}_i^p, \mathbf{t}_i^d, \mathbf{t}_i^{\dot{p}}, \vec{t}_i^{v}\}$.
\end{itemize}
The first input to the neural network is the robot state $\mathbf{x}_i$, where $\vec{r}_i \in \mathbb{R}^3$ and $\dot{\vec{r}_i} \in \mathbb{R}^3$ are robot's position and velocity, $\vec{w}_i \in \mathbb{R}^6$ is robot's orientation, and $\vec{e}_i \in \mathbb{R}^{12}$, $\dot{\vec{e}_i} \in \mathbb{R}^{12}$ are end-effectors' positions and velocities.
The second input is a surrounding root trajectory with $K$ horizon steps, where the half are for history and the rest are for desired future. Here, $\vec{t}_i^p \in \mathbb{R}^{2 K}$ are planar position trajectory, $\vec{t}_i^d \in \mathbb{R}_i^{2 K}$ are planar  direction trajectory,  $\vec{t}^{\dot{p}} \in \mathbb{R}^{2 K}$ are planar  velocity trajectory, and $\vec{t}_i^{v} \in \mathbb{R}^{K}$ are desired speed. We use K=12 in our work which represents a two-second trajectory. 
%\jhc{We additionally add 21 sensors to sense obstacles within 1m range parallel to the forward facing direction to react accordingly to obstacles.} %automatic jumping on obstacles.}

The output of the planner are the next robot state $\mathbf{x}_{i+1}$ and the predicted future trajectory  $\mathbf{t'}_{i+1} =\{\mathbf{t'}_{i+1}^p, \mathbf{t'}_{i+1}^d, \mathbf{t'}_{i+1}^{\dot{p}}\}$
where $\mathbf{t'}_{i+1}^p \in \mathbb{R}^{K}$, $\mathbf{t'}_{i+1}^d \in \mathbb{R}^{K}$, $\mathbf{t'}_{i+1}^{\dot{p}} \in \mathbb{R}^{K}$. 
All features of the input and output are designed in the robot's local coordinate at the $i$-th time step.

During planning, the motion planner takes the current state and surrounding root trajectory to generate the next state and predicted root trajectory. The output is then worked as the input for the next time step. Different motions can be generated by varying the future root trajectory according to the high-level command with an autoregressive blending process. Overall, the control diagram can be best seen in Fig.~\ref{fig: Overview}.

\subsection{Target Trajectory Blending}
\label{sec:trajectory blending}
When providing the run-time input for the motion planner, the future root trajectory $\mathbf{t}_i^p, \mathbf{t}_i^d$ is generated by blending the predicted future trajectory from the motion planner and a target trajectory from the high-level command to create a desired future trajectory for the next input. Here, we utilize trajectory blending technique frequently used in kinematic animation \cite{holden2017phase, zhang2018mode}:
% \begin{equation}
%     \text{TrajectoryBlend}(\vec{a}_0, \vec{a}_1, t, \sigma) = (1 - t^\sigma) \vec{a_0} + t^\sigma \vec{a_1},
% \end{equation}

\begin{equation}
    \vec{t}_i^p(z) = (1- z^{\sigma_p}) \vec{t'}_i^p(z) + t^{\sigma_p} \vec{\hat{t}}_i^p(z), z \in [0,1]
\end{equation}
where $\vec{t}_i^p(z)$ denotes desired trajectory's planar position $z$ seconds after $i$-th timestep, and $\vec{t'}_i^p$ and $\vec{\hat{t}}_i^p$ corresponds to that of predicted and target trajectories.
The values $\vec{t'}_i^p(z)$ at future time $z$ is obtained by linearly interpolating neural-net generated predicted trajectory.
Target trajectory $\vec{\hat{t}}_i^p(z)$ is manipulated by user commands such as target linear and angular velocities.
Here, $\sigma_p > 0$ defines responsiveness with respect to user command.
Trajectory blending for other features such as planar direction $\vec{t}_i^d (z)$ and velocity $\vec{t}_i^{\dot{p}}(z)$ are done in a similar fashion.

% Choose correct place to put.
This target trajectory method enables fluent integration with navigation frameworks~\cite{habitat19iccv,szot2021habitat}, where navigation restrictions due to walls or obstacles are represented by navigation mesh~\cite{van2016comparative}.
In complex indoor scenes, we generate a navigation mesh to constrain our target trajectory to the navigable area before blending it with the future predicted trajectory for generating the desired future trajectory which is used as an input for the autoregressive planner.

Our target trajectory is constrained inside the navigation mesh either by sliding or stopping at the boundary of the navigable area. Even though this constraining process is done before the blending process, the desired future trajectory tends to remain inside this navigable area since all input target trajectories are hard constrained to this area.

This autoregressive method ensures the motion planner generates physically feasible motions while following the high-level command. We will demonstrate it in Section \ref{sec: result}.

\section{Experiments and Results}
% \ty{The entire section is very draft}
\label{sec: result}

% \begin{figure*}[ht]
% \centerline{\includegraphics[width=1.0\linewidth]{figures/Figure_results.png}}
% \caption{\small{Snapshots of our approach in action on the Aliengo in simulation.}}
% \label{fig: indoor}
% \vspace{-10pt}
% \end{figure*}

We now present our experimental analysis of our Auto-Regressive Motion Planner, ARMP. Our experiments are designed to answer the following questions:
\begin{enumerate}
    \item Can ARMP generate trajectories in various scenarios?
    \item Are ARMP-generated trajectories physically feasible?
    \item Can ARMP be employed as a low-level controller for indoor navigation?
\end{enumerate}
For the rest of the section, we will first provide information about our experimental setting and then discuss each research question in the corresponding subsections.
% The following subsections are constructed as follows: first, we introduce the basic information about our learning and experiment setting. Then we test if ARMP is able to generate diverse motions to answer Question (1). For Question (2), we conducted an experiment to validate the physical feasibility of the generated motion. Lastly, we apply our approach AWMP for indoor navigation tasks to answer Question (3).

% \ty{Three questions: (1) are generated motions expressive enough to handle different terrains? (2) are generated motion physically feasible? (3) can ARMP be applied to indoor navigation task?}

\subsection{Experiment Setting}
% \ty{Here we show what simulator we are using. Robot? Training details, NN structure, training time, GPU.... }
We implement our framework using PyTorch~\cite{paszke2019pytorch}. 
Our motion planner is constructed using the mixture-of-expert architecture with 4 experts as suggested in the work of Zhang et al.~\cite{zhang2018mode}. Each expert is a multi-layer perceptron with 2 hidden layers of size [512, 512] with ELU\cite{clevert2015fast} activations. For the motion library, we construct the majority of the motion library via trajectory optimization, where we generate 56 trajectories with durations ranging from 20 to 65 seconds. The trajectories are optimized by changing target position, orientation, time horizon, and gaiting frequencies, resulting in total of 43 minutes of data with about 158000 state transitions when discretized into 30 frames per second. We generate a jumping motion by retargeting the captured dog's jumping motion in Zheng et al.~\cite{zhang2018mode}. The single jumping trajectory of 9 seconds length is duplicated 11 times and then added to the motion library. The motion library is then used for training our module via Google Colab\cite{bisong2019google}. We use AlienGo~\cite{unitree} from Unitree as our robotic platform and use PyBullet~\cite{coumans2016pybullet} only for visualization.

\subsection{Generating Trajectories in Various Scenarios} %first two questions

\begin{figure}
     \centering
     \begin{subfigure}[b]{0.25\textwidth}
         \centerline{\includegraphics[width=1.1\linewidth]{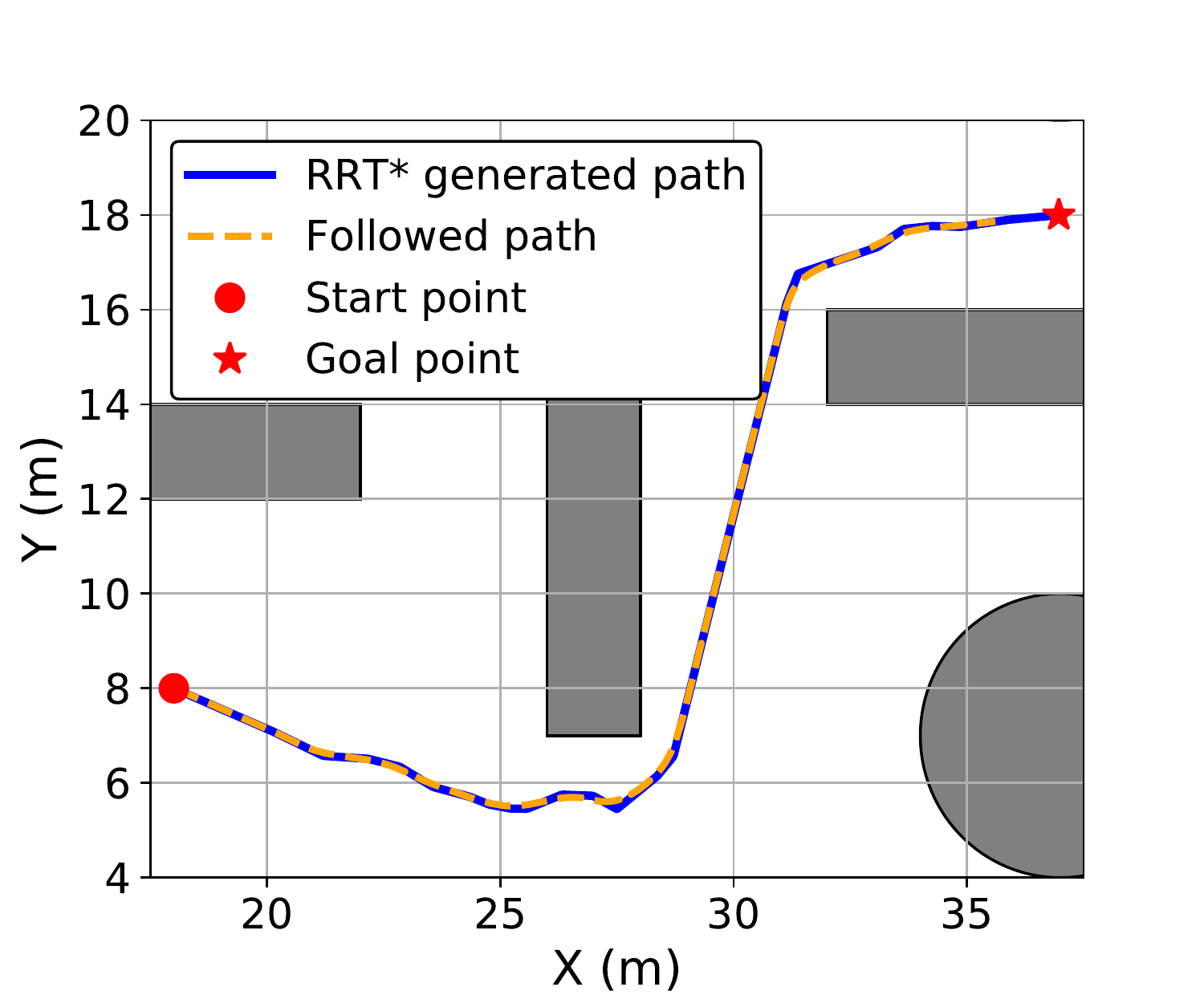}}
         \caption{Path Following}
         \label{fig:follow1}
     \end{subfigure}%
     \begin{subfigure}[b]{0.25\textwidth}
         % \vspace{-10pt}
         \centerline{\includegraphics[width=1.1\linewidth]{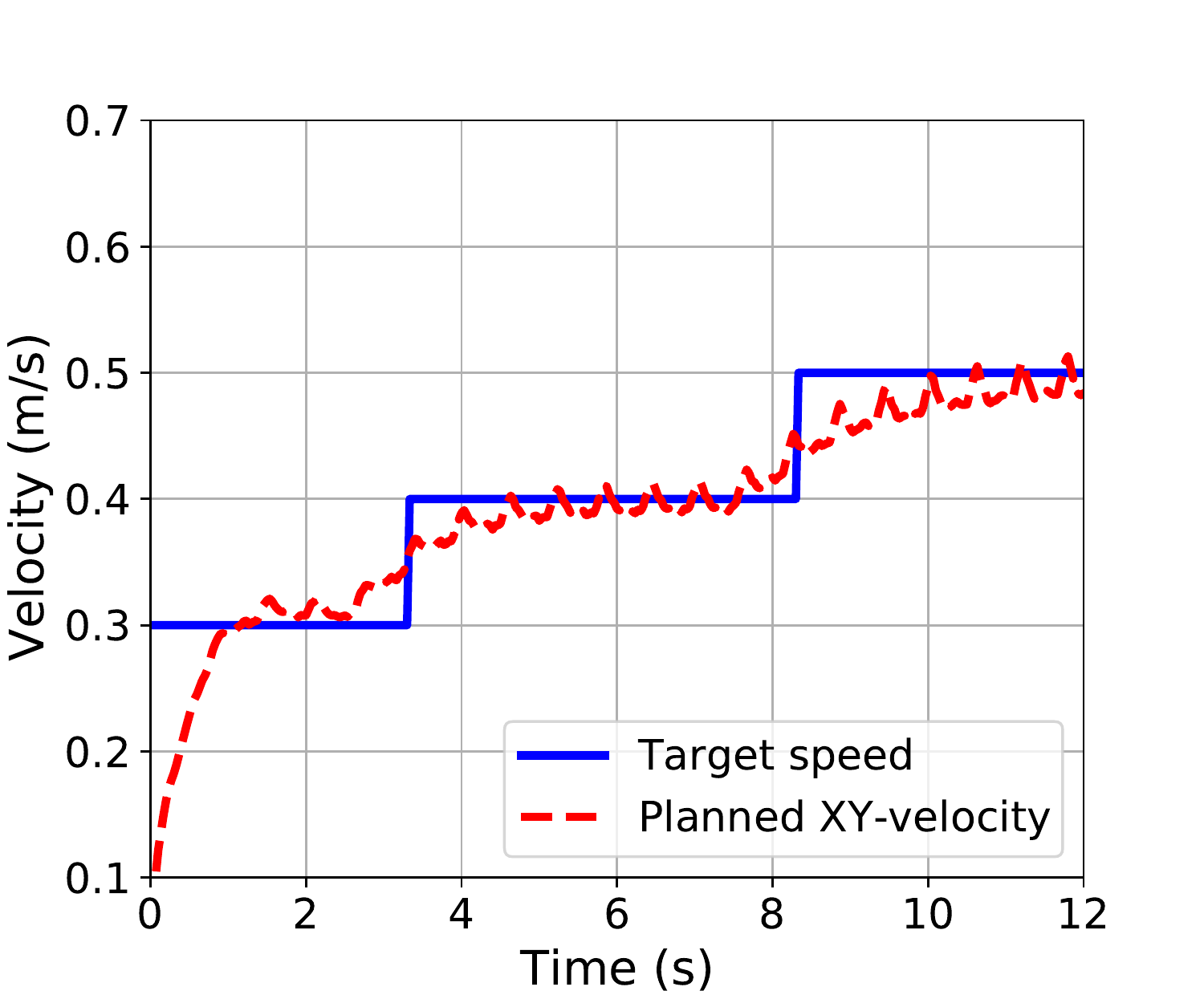}}
         \caption{Velocity Tracking}
         \label{fig:follow2}
     \end{subfigure}\hfill%
\caption{\small{Tracking results of the given COM path (\textbf{Left}) and the velocity profile (\textbf{Right}).}}
\label{fig:follow}
\end{figure}

\begin{figure}
     \centering
     \begin{subfigure}[b]{0.5\textwidth}
        \centerline{\includegraphics[width=1.1\linewidth]{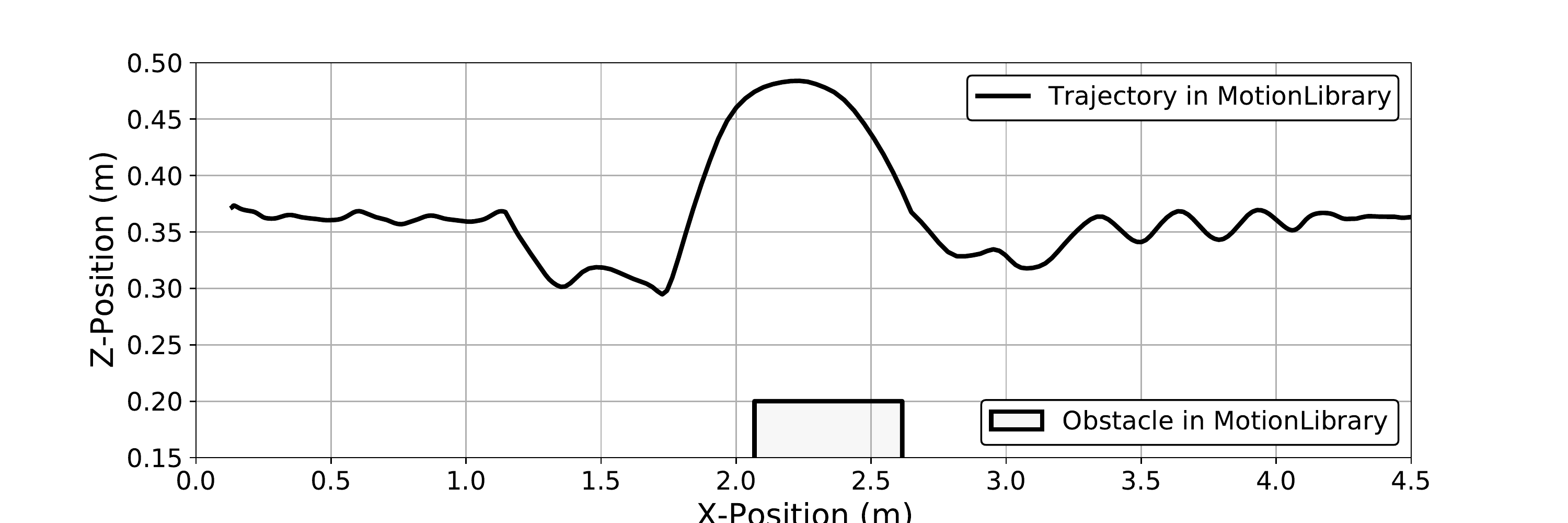}}
         % \vspace{-5pt}
         \caption{Jump Trajectory in Motion Library}
         \label{fig:jump1}
     \end{subfigure}\hfill
     \begin{subfigure}[b]{0.5\textwidth}
         % \vspace{-5pt}
         \centerline{\includegraphics[width=1.1\linewidth]{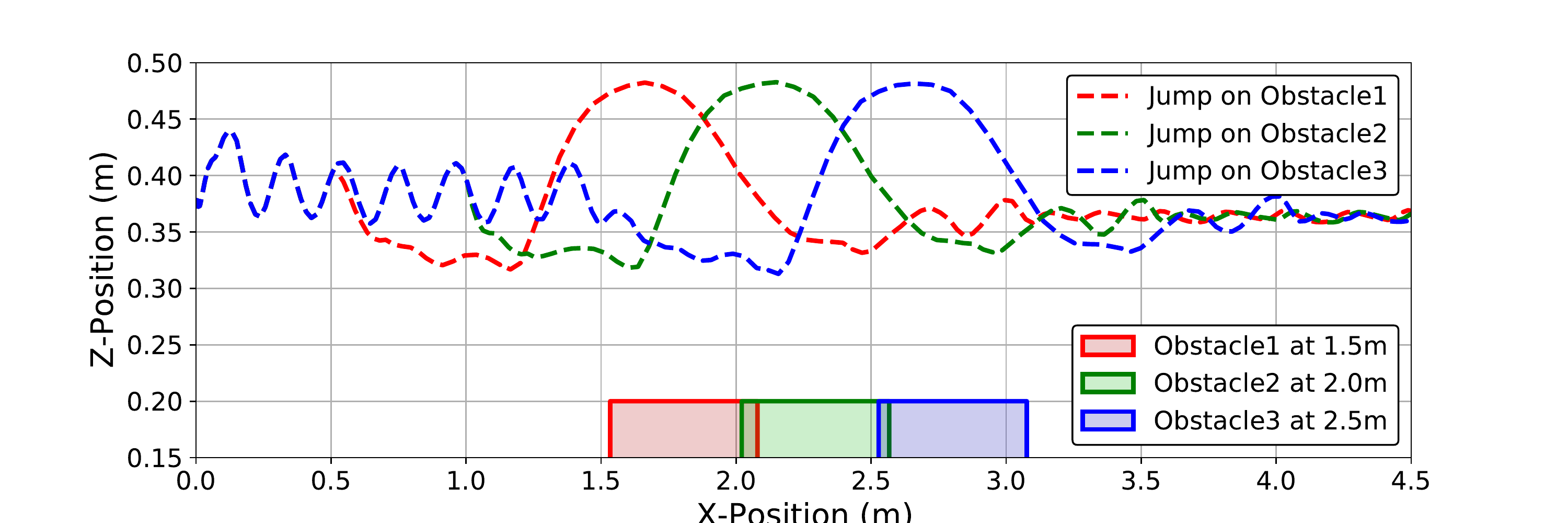}}
         % \vspace{-5pt}
         \caption{Generated Jump Trajectories}
         \label{fig:jump2}
     \end{subfigure}\hfill%
\caption{\small{Top plot represents the jumping trajectory used in our motion library. Bottom plot shows trajectories generated by our framework with obstacles placed in different positions. Smooth transition between trotting and  the jumping motion can be observed.}}
\label{fig:jump}
\end{figure}

% Figure 2. COM and Velocity tracking
To demonstrate that our framework can generate diverse trajectories, we first check the effectiveness of our framework on following the given desired center of mass (COM) path. 
First, we generate a motion trajectory to follow a path to navigate in cluttered environment, which is generated using RRT*~\cite{karaman2011sampling}. Please note that path generation or viaual navigation is not the main focus of the paper. Rather, we aim to evaluate the capability for generating motion plans with respect to the given high-level task.
We can see in Fig. \ref{fig:follow1} that our generated trajectories can smoothly follow the desired COM path with the maximum deviation of $15.1$~cm.

We also evaluate the velocity following performance of ARMP by commanding a manually-designed velocity profile (Fig. \ref{fig:follow2}), which shows the maximum velocity error of $0.065$~m/s.
We observe the smoothed tracking of velocity due to target trajectory blending discussed in \ref{sec:trajectory blending}. Also note that the given discrete velocity profile is impossible for quadrupedal robots that needs more time for acceleration.

% Figure 3. Jumping trajectories.
We also show how the robot jumps over the obstacle at different distances by plotting the COM trajectories in Fig.~\ref{fig:jump}. It shows how ARMP adjusts and interpolate the existing plans in the motion library to jump over obstacles at different locations.
Since the motion library consists of a single jumping motion as shown in Fig.~\ref{fig:jump1}, we see that the planner retrieves the same motion from the library at the based on terrain condition.
Although the motion library did not contain any motion trajectory that transitions between jumping and trotting, our framework smoothly blends the two based on current state and terrain condition.

\subsection{Physical Feasibility of the Generated Plans}

\begin{figure}
     \centering
     \begin{subfigure}[b]{0.5\textwidth}
         \centerline{\includegraphics[width=1.1\linewidth]{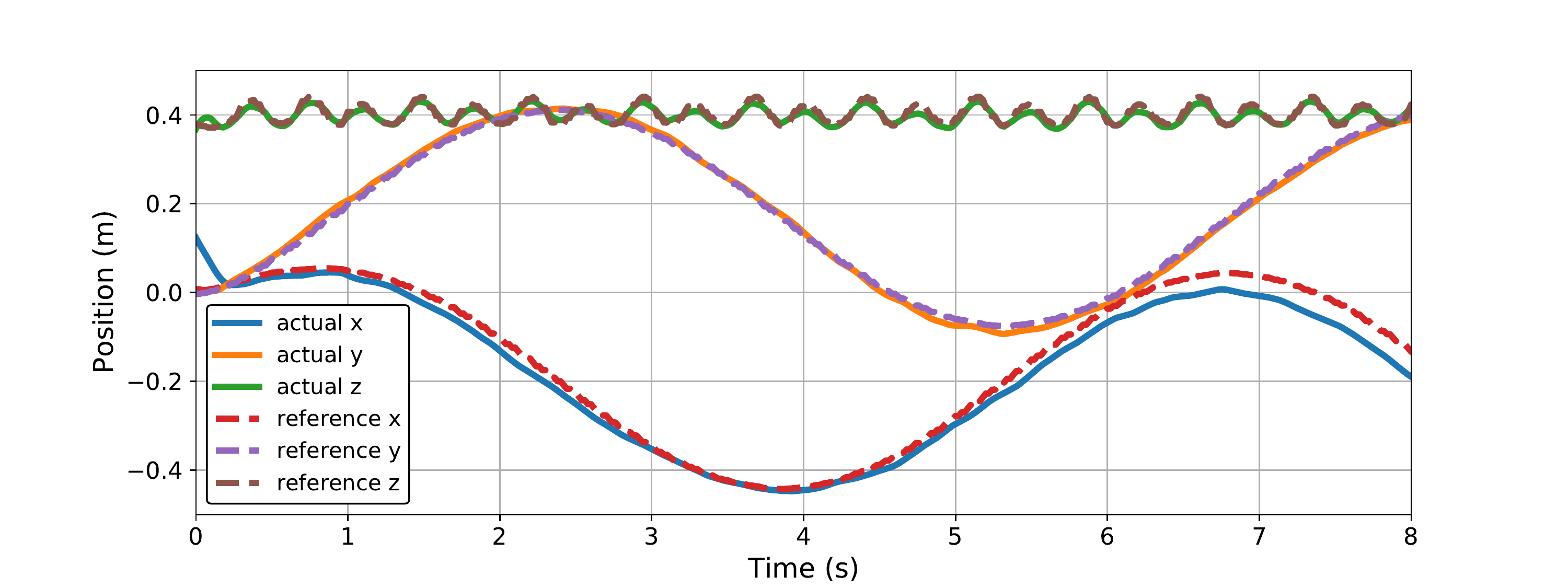}}
         \caption{Base position (m)}
         \label{fig:track turn 0}
     \end{subfigure}\hfill%
     \begin{subfigure}[b]{0.5\textwidth}
         \centerline{\includegraphics[width=1.1\linewidth]{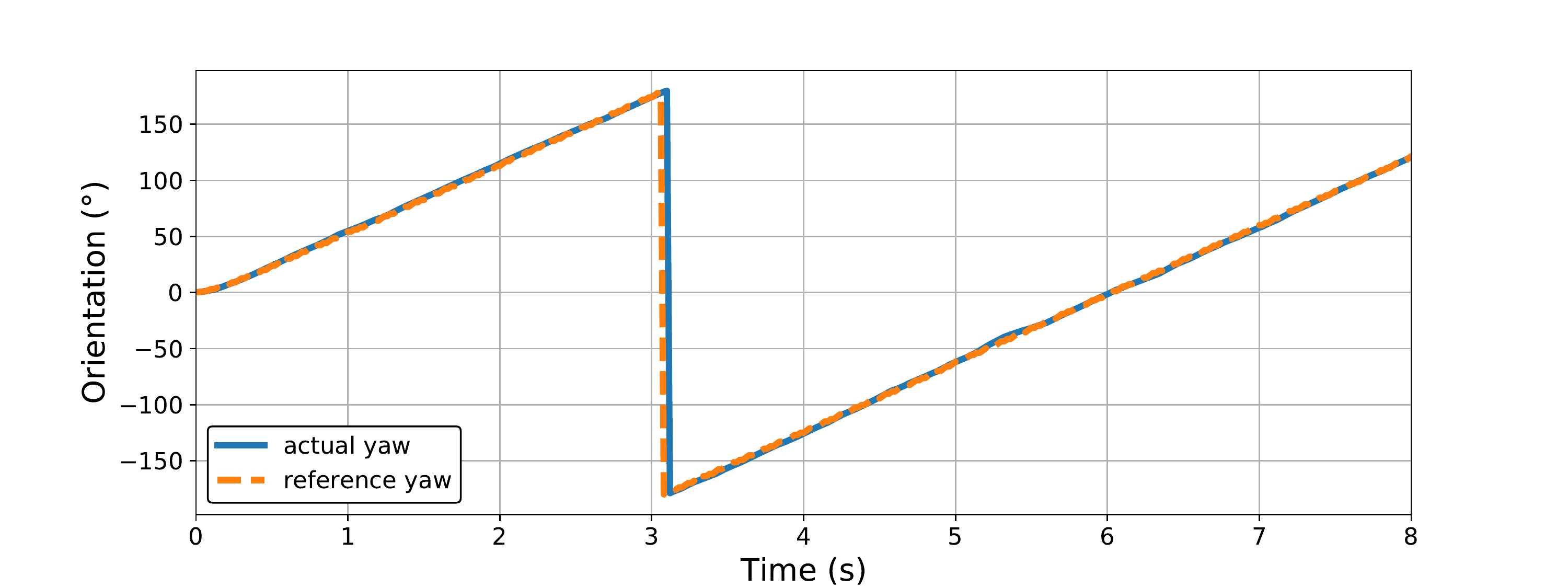}}
         \caption{Base orientation (°)}
         \label{fig:three sin x}
     \end{subfigure}\hfill%
     \begin{subfigure}[b]{0.5\textwidth}
         \centerline{\includegraphics[width=1.1\linewidth]{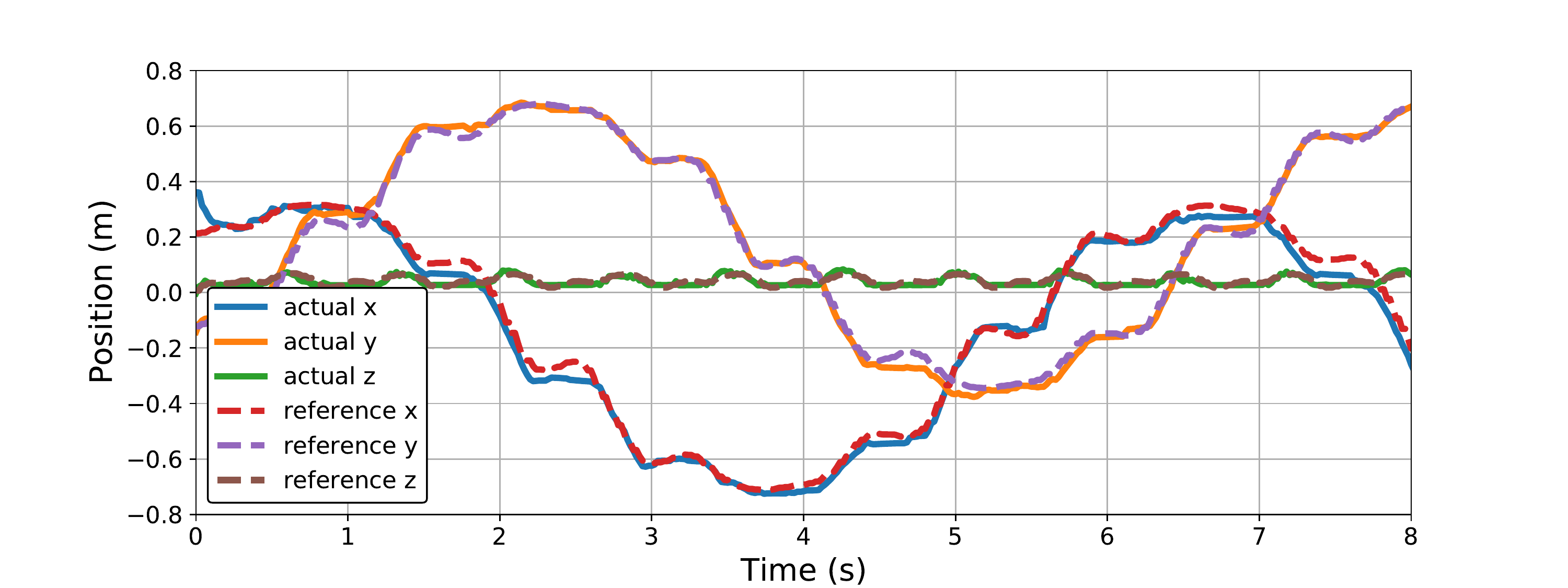}}
         \caption{Front right toe position (m) }
         \label{fig:three sin x}
     \end{subfigure}\hfill%
     % \begin{subfigure}[b]{0.25\textwidth}
     %     \centering
     %     \includegraphics[width=\textwidth]{figures/Track_turn_3.pdf}
     %     \caption{Front Left Toe position}
     % \end{subfigure}
\caption{\small{Validation in a full-scale simulation by training a tracking policy. Our planned trajectory is shown in dashed line and the tracked trajectory is in bold line.}}
\label{fig:tracking}
\end{figure}

% Motivation: why is it hard to evaluate the feasibility using optimization?
While ARMP can generate various motions by interpolating the existing ones, the physical feasibility of the generated motions must be validated to be used ARMP as a motion planner.
% However, it is not straightforward to validate the physical feasibility of the given motion plan.
Intuitively, we may want to evaluate the plan by evaluating the cost and constraint functions of the trajectory optimization described in Section~\ref{sec:data_generation}, but the comparison is not straightforward because ARMP and the trajectory optimization use different state representations: a list of vectors and Hermite polynomials. We also do not directly compare the ARMP and trajectory optimization results for the same task because two planners can generate feasible but different motion plans, particularly when it comes to long-horizon tasks.

% Figure 4. DeepMimic analysis
Instead, we check the physical feasibility by leveraging a physics-based simulation. Particularly, we utilize a neural motion imitation policy to track the given motion plan. If it can closely follow the motion plan, it indicates that it is feasible in a high-frequency, full-scale physics-based simulation. We use a deep reinforcement learning algorithm, Proximal Policy Optimization~\cite{schulman2017proximal}, to train such a tracking policy using the motion imitation framework described in Peng et al.~\cite{peng2018deepmimic}. For more details, please refer to the original paper. Fig.~\ref{fig:tracking} shows the ARMP motion plan and the trajectory in tracked in physics-based simulation~\cite{raisim} for a $360^\circ$ circular turning task, which is unseen during training. The tracking policy was able to closely follow the given COM and end-effector trajectories, which indicates the physical feasibility of the planned motion.

\subsection{Application: Indoor Navigation} % last question
\label{sec: indoor navigation}

\begin{figure*}
    \centering
    \setlength{\tabcolsep}{1pt}
    \renewcommand{\arraystretch}{0.7}
    \begin{tabular}{c c c c c}
    \includegraphics[width=0.185\textwidth]{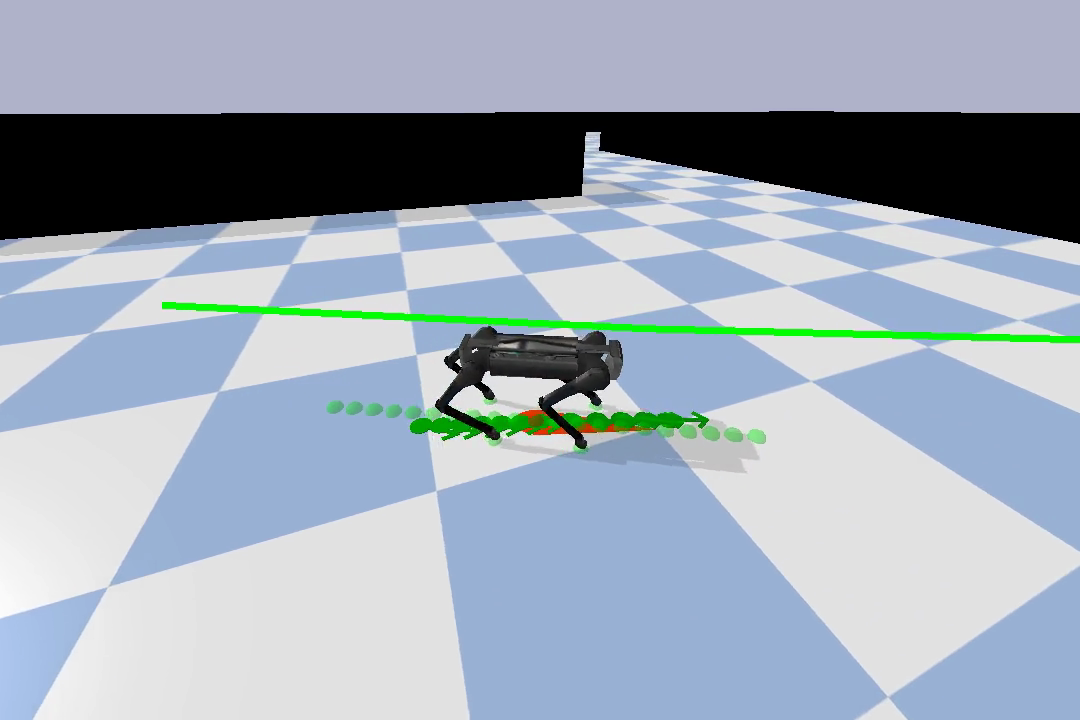} &
    \includegraphics[width=0.185\textwidth]{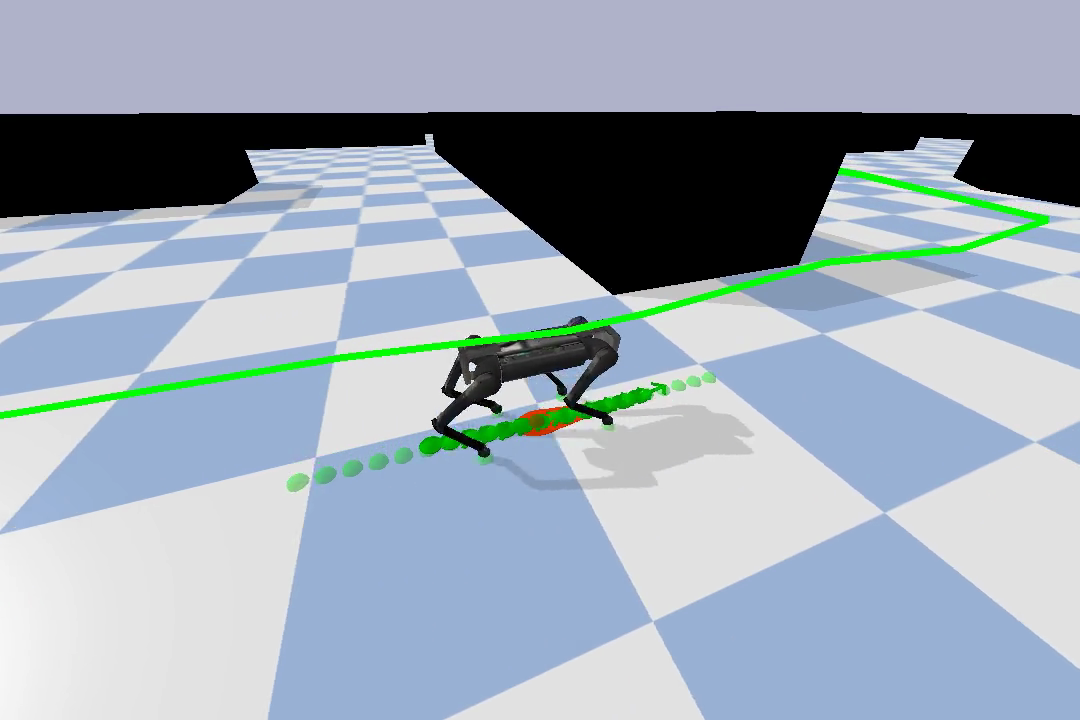} &
    \includegraphics[width=0.185\textwidth]{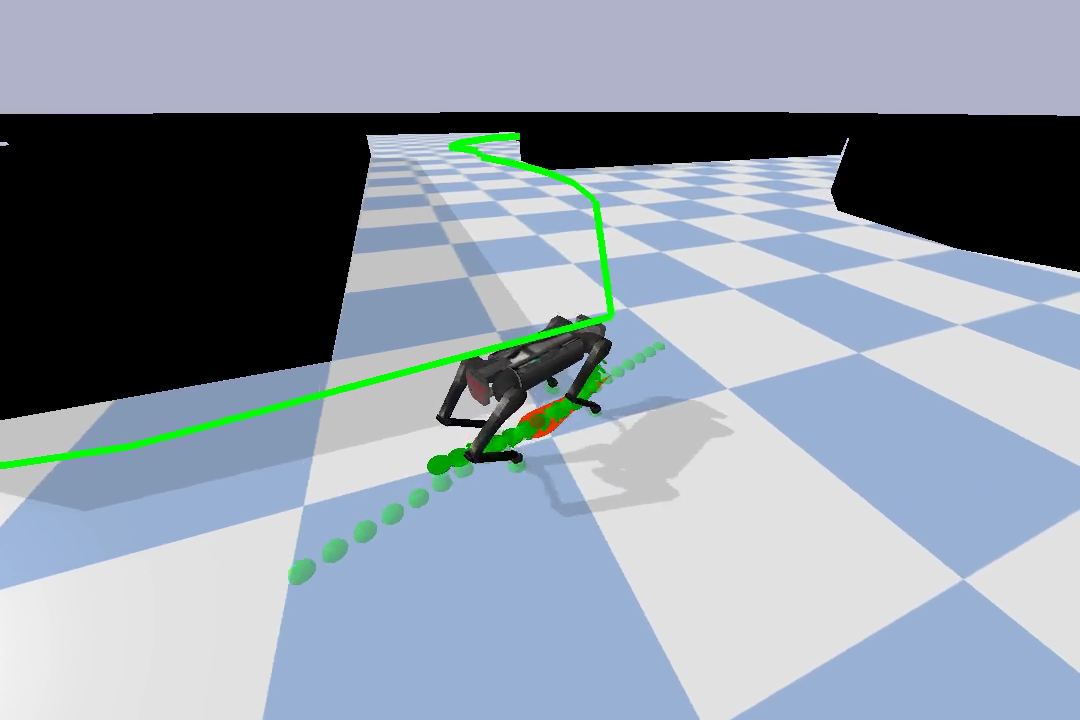} &
    \includegraphics[width=0.185\textwidth]{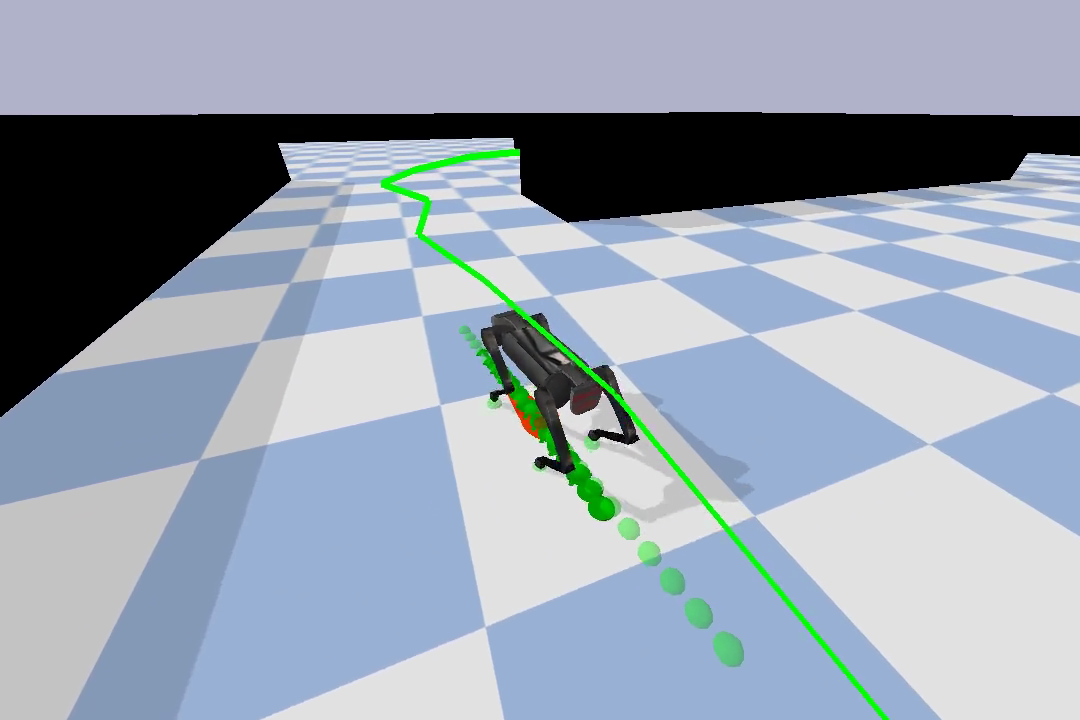} &
    \includegraphics[width=0.185\textwidth]{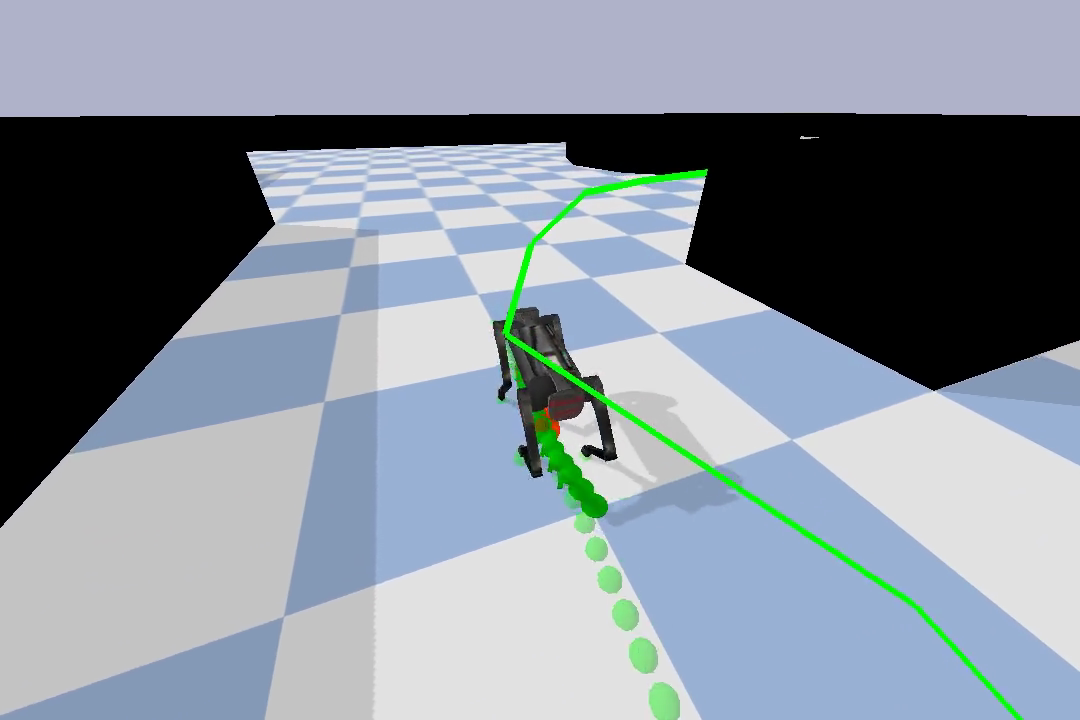} \\
    \includegraphics[width=0.185\textwidth]{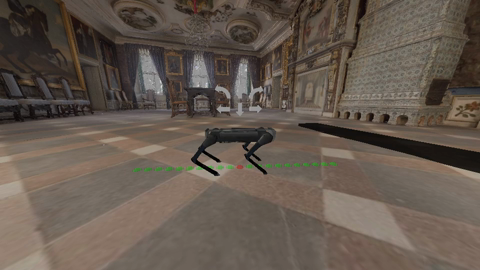} &
    \includegraphics[width=0.185\textwidth]{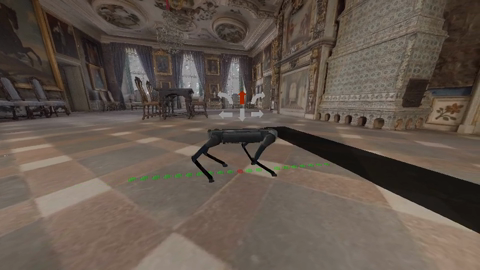} &
    \includegraphics[width=0.185\textwidth]{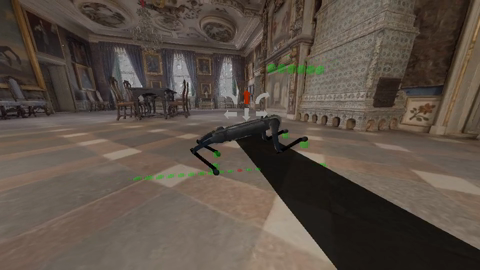} &
    \includegraphics[width=0.185\textwidth]{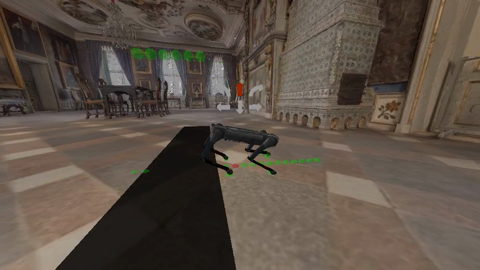} &
    \includegraphics[width=0.185\textwidth]{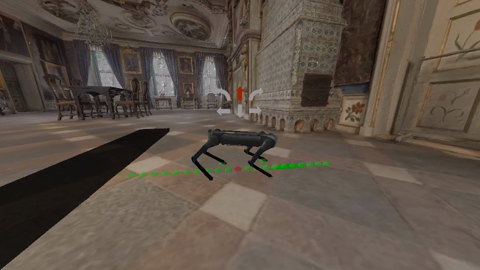} \\ 
    \includegraphics[width=0.185\textwidth]{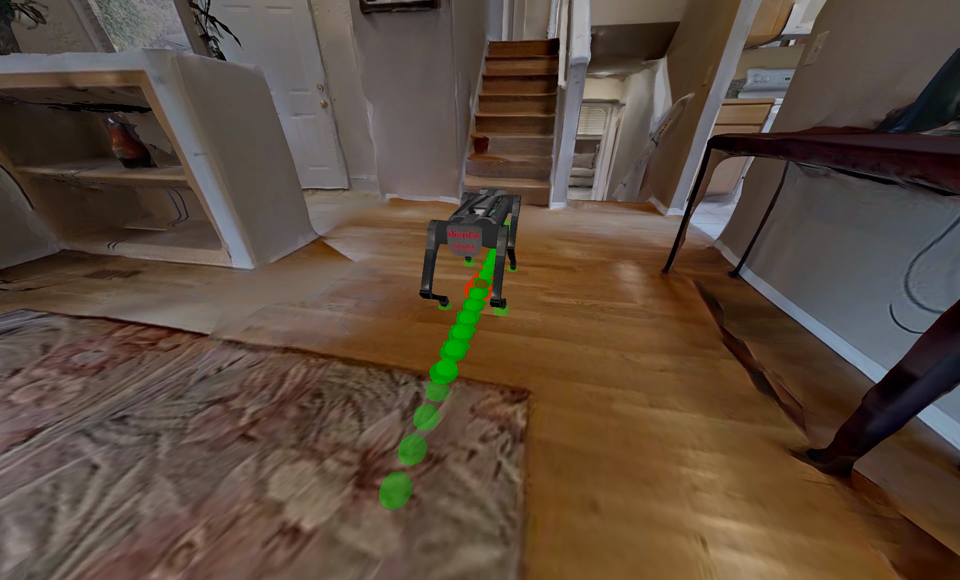} &
    \includegraphics[width=0.185\textwidth]{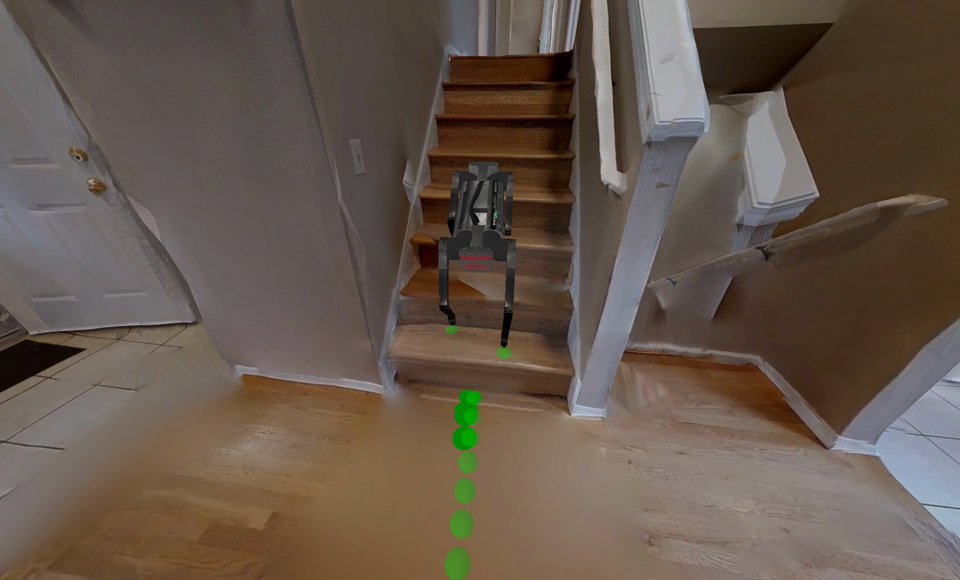} &
    \includegraphics[width=0.185\textwidth]{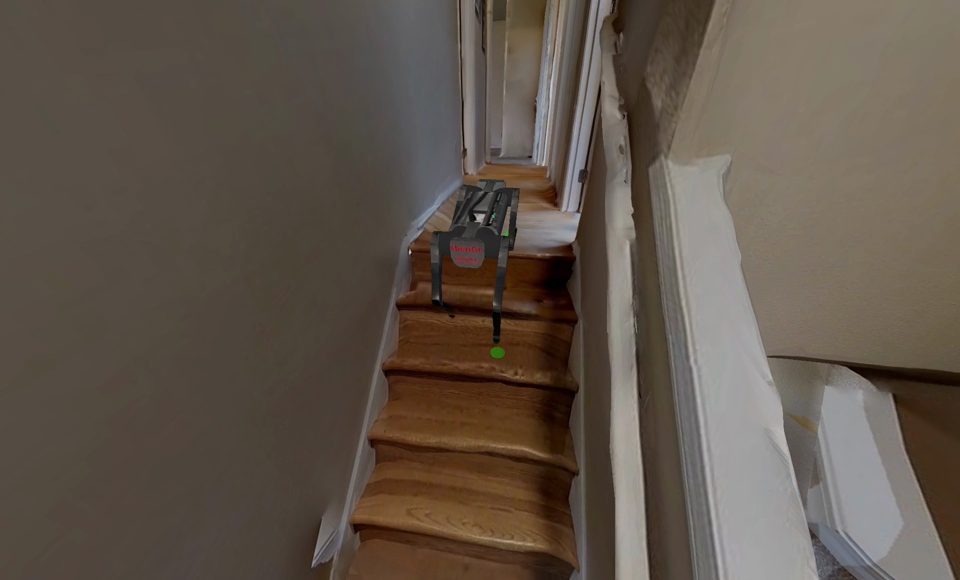} &
    \includegraphics[width=0.185\textwidth]{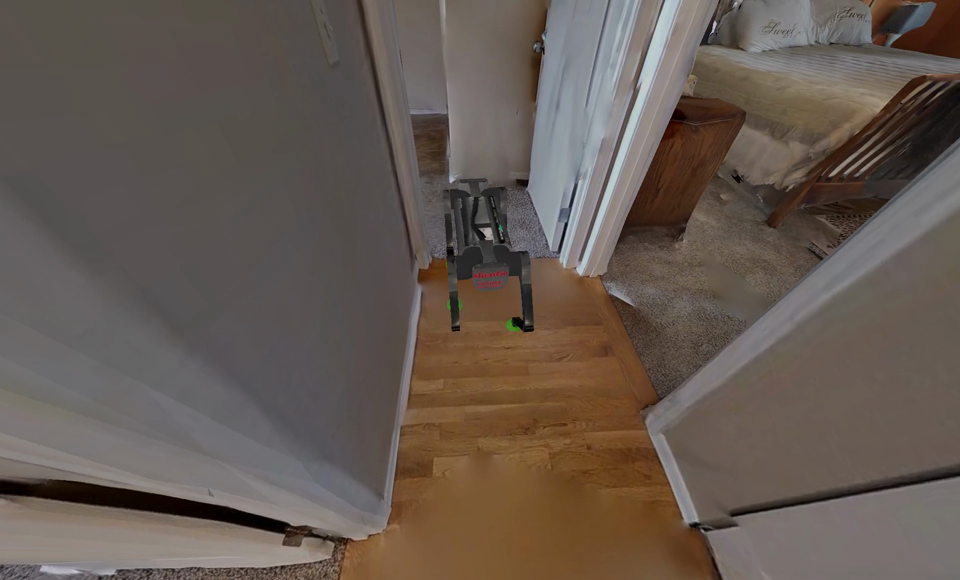} &
    \includegraphics[width=0.185\textwidth]{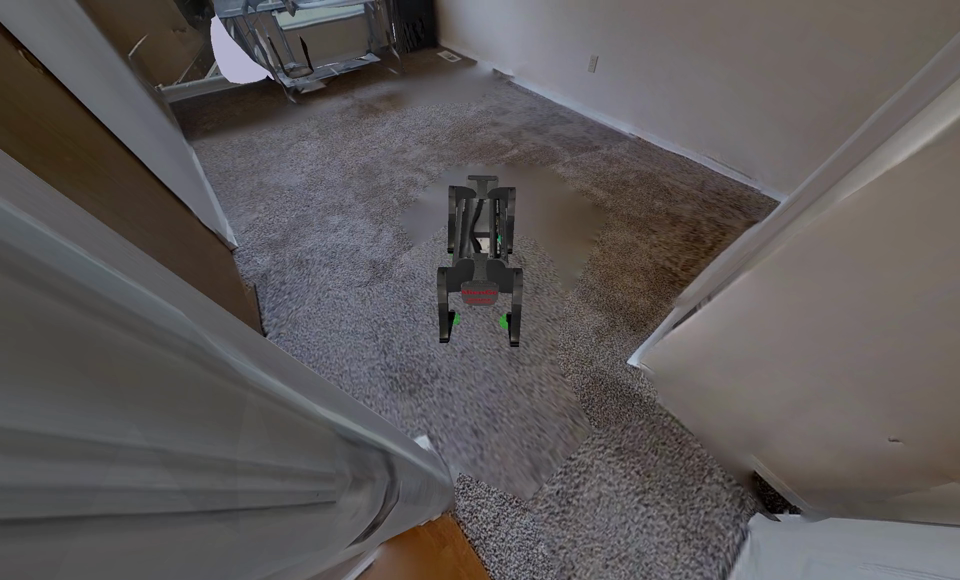}   
    \end{tabular}
    
    \caption{Generated motions for indoor navigation.
    \textbf{Top:} a robot follows the path between obstacles.
    \textbf{Middle:} a robot jump over the obstacle.
    \textbf{Bottom:} a robot walks up the stair to get to the room on the second floor.
    }
    \label{fig: indoor}
\end{figure*}

% Motivation
% This section aims to answer the question of whether ARMP can be used as a low-level controller for indoor navigation. 
Many works on indoor navigation frameworks~\cite{habitat19iccv, szot2021habitat} only address the problem of reaching the desired goal position without climbing stairs or encountering negotiable obstacles due to the nature of wheel-based robots. On the other hand, our results show that ARMP can utilize the full potential of the legged robot, as ARMP can generate versatile motions for different terrains. This makes ARMP an effective low-level motion generator for indoor navigation with complex terrain settings. For all the experiments, high-level trajectories are generated manually or using the off-the-shelf motion planner such as RRT* because learning a navigation policy is not the main focus of the paper, although it can be approached by other visual navigation techniques~\cite{kareer2022vinl}.

In this section, we investigate ARMP in the following three scenarios. All the motions can be seen in Fig.~\ref{fig: indoor} and the supplemental video. 

\subsubsection{Simple Navigation}
One of the biggest challenges of indoor navigation is that it consists of various unnavigable areas like walls and obstacles. The goal is to move in a clustered environment to a goal located (19m, 10m) relative to robot’s start position. In this task, we use RRT* to generate a  collision free path of the robot’s base. ARMP uses this path as a target trajectory to be blended, and plans a full body trajectory that reaches the goal in 60 seconds.

\subsubsection{Obstacle Navigation}
In the second scenario, we consider a scene with an obstacle with the dimension of $0.2$~m height and $0.4$~m width. With this obstacle located in the path of robot’s target trajectory, ARMP generates a compelling full-body trajectory that includes a motion to jump over the obstacle without collision. Please note that typical navigation formulation for wheeled robots cannot follow such path.

\subsubsection{Stair Navigation}
Finally, we use ARMP to generate a motion plan which starts on the first floor and ends on the second floor. The stair in the scene is 21~cm height for one step, and there are 8 steps in total. We demonstrate that ARMP can plan a total 14.38~m path that leads the robot to the upper floor, which includes a reasonable stair-climbing motion, where the height difference is compensated using inverse kinematics. Once again, we would like to stress that two-story navigation is not thoroughly investigated in the navigation community due to the assumption of simple wheeled robots. We hope that the proposed ARMP can be used for studying the full navigation capability of legged robots, including multi-story buildings and cluttered scenes.

\section{Conclusion and Future Work}
We present a framework for learning an auto-regressive motion planner (ARMP) to generate physically plausible motion plans for various tasks and environments. We describe how we construct a motion library using a trajectory optimization method. We then illustrate how the motion planner generates physically feasible trajectories via deep learning while following high-level commands using an autoregressive manner. Our evaluation results show that the proposed ARMP can produce a variety of physically feasible motions, including walking, turning, jumping, and stair climbing. Finally, we show that the learned motion planner can be used for complex indoor navigation tasks, such as navigating to goals on different floors or goals behind an obstacle that can not go around, which is not impossible for wheeled robots.

While our work shows promising results, it is still challenging to collect the required trajectories in the motion library due to the difficulty of tasks and the complexity of environments. For instance, it is not straightforward to solve a jumping motion using trajectory optimization, which makes us retarget the captured motion of a real dog. In another example, the environment can be too complex to consider all the possible scenarios, such as stepstones with various gaps or densely populated obstacles. We can approach these challenges by developing a better low-level trajectory optimization module or adaptively sampling the environments based on the learning progress.

If the motion library becomes larger and larger, our learning-based motion planner may not be possible to learn all the trajectories in the database. In this case, it may generate inaccurate trajectories with poor physics or diverged behaviors. One possible solution is to increase the size of the network or to introduce a new architecture that has better consideration of spatial and temporal relationships between frames. We will leave this to our future work.

\section*{Acknowledgement}

This work is supported by MORAI, Inc.

\bibliographystyle{IEEEtran} 
\bibliography{00.bibliography.bib.tex}

\end{document}